\documentclass[lettersize,journal]{IEEEtran}
\usepackage{amsmath,amsfonts}
\usepackage{algorithmic}
\usepackage{algorithm}
\usepackage{array}
\usepackage{textcomp}
\usepackage{stfloats}
\usepackage{verbatim}
\usepackage{graphicx}
\usepackage{cite}

\usepackage{graphicx}
\usepackage{graphicx}
\usepackage{microtype}      
\usepackage{xcolor}         
\usepackage{amsmath}
\usepackage{multirow}
\usepackage{multicol}
\usepackage{adjustbox}
\usepackage{pifont}
\usepackage[utf8]{inputenc}
\usepackage[T1]{fontenc}   
\usepackage{booktabs}     
\usepackage{amsfonts}      
\usepackage{nicefrac}  
\usepackage{enumitem}
\newcommand{\xmark}{\ding{55}}
\newcommand{\vmark}{\ding{51}}
\usepackage{tikz}
\usepackage{caption}

\usepackage{subcaption}
\usepackage[table,xcdraw]{xcolor}
\usepackage[normalem]{ulem}
\useunder{\uline}{\ul}{}
\usepackage[colorlinks=true, linkcolor=blue, citecolor=blue, urlcolor=blue]{hyperref}
\begin{document}

\title{Measuring the Measurers: Quality Evaluation of Hallucination Benchmarks for Large Vision-Language Models}

\author{Bei Yan, Jie Zhang,~\IEEEmembership{Member,~IEEE}, Zheng Yuan, Shiguang Shan,~\IEEEmembership{Fellow,~IEEE}, Xilin Chen,~\IEEEmembership{Fellow,~IEEE}
\thanks{Bei Yan, Jie Zhang, Zheng Yuan, Shiguang Shan, Xilin Chen are with State Key Laboratory of AI Safety, Institute of Computing Technology, Chinese Academy of Sciences, and also with University of Chinese Academy of Sciences. E-mails: \{yanbei23s, zhangjie\}@ict.ac.cn, zheng.yuan@vipl.ict.ac.cn, 
\{sgshan, xlchen\}@ict.ac.cn.}
}

\maketitle

\begin{abstract}

Despite the outstanding performance in multimodal tasks, Large Vision-Language Models (LVLMs) have been plagued by the issue of hallucination, i.e., generating content that is inconsistent with the corresponding visual inputs. While previous works have proposed various benchmarks to evaluate this issue, the quality of these evaluations remains unverified. We observe that some of these benchmarks may produce inconsistent evaluation results across repeated tests or fail to align with human evaluation.
To address this, we propose a \textbf{H}allucination benchmark \textbf{Q}uality \textbf{M}easurement framework (\textbf{HQM}), which leverages specific indicators to assess both reliability and validity. Our empirical analysis using HQM reveals and pinpoints potential evaluation issues in existing benchmarks, exposing a critical gap in current hallucination evaluation. To bridge this gap, we propose HQH\footnote{\url{https://github.com/HQHBench/HQHBench}}, a \textbf{H}igh-\textbf{Q}uality \textbf{H}allucination benchmark, which demonstrates superior reliability and validity under HQM, serving as a credible evaluation tool. Our large-scale evaluation of popular LVLMs on HQH reveals severe hallucination problems, which occur not only in the models' main answer to a question but also in additional analysis. This highlights the necessity for future model improvements to effectively mitigate hallucinations and reduce the associated security risks in real-world applications.

\end{abstract}


\section{Introduction}

\IEEEPARstart{R}{ecently}, the emergence of Large Language Models (LLMs) has led to a great revolution in the field of artificial intelligence. Building on the success of LLMs, Large Vision-Language Models (LVLMs) have made remarkable advancements. These models usually use LLMs as the foundational architecture and align features from other modalities accordingly, demonstrating exceptional capabilities across various multimodal tasks, such as image captioning and visual question answering (VQA). Despite their outstanding performance, LVLMs are significantly plagued by the issue of hallucination, i.e., generating content that appears plausible but contradicts the visual input~\cite{bai2024hallucinationsurvey,liu2024survey, zhang2024b}. Hallucinations not only adversely affect model performance in specific tasks but also pose serious security risks in real-world applications, particularly in sensitive fields like medicine~\cite{peng2023medical} and law~\cite{lai2024law}, where users lacking domain expertise may over-rely on the models, leading to potentially harmful consequences.

\begin{figure}[t!]
  \centering
\includegraphics[width=\linewidth]{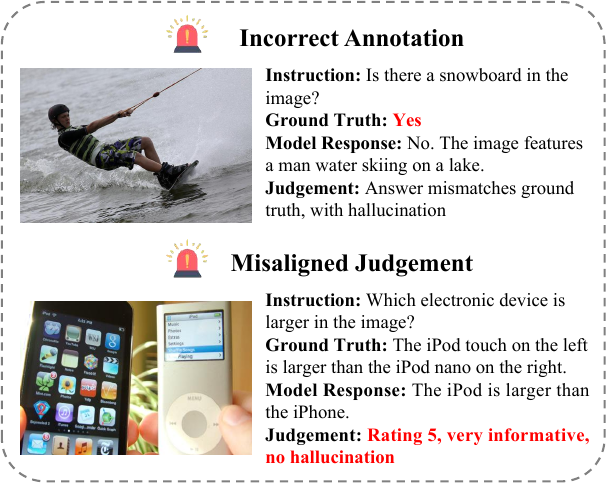}
  \caption{Examples of potential evaluation issues in current hallucination benchmarks.}
  \label{issue_examples}
\end{figure}

\begin{figure*}
\centering
 \includegraphics[width=\linewidth]{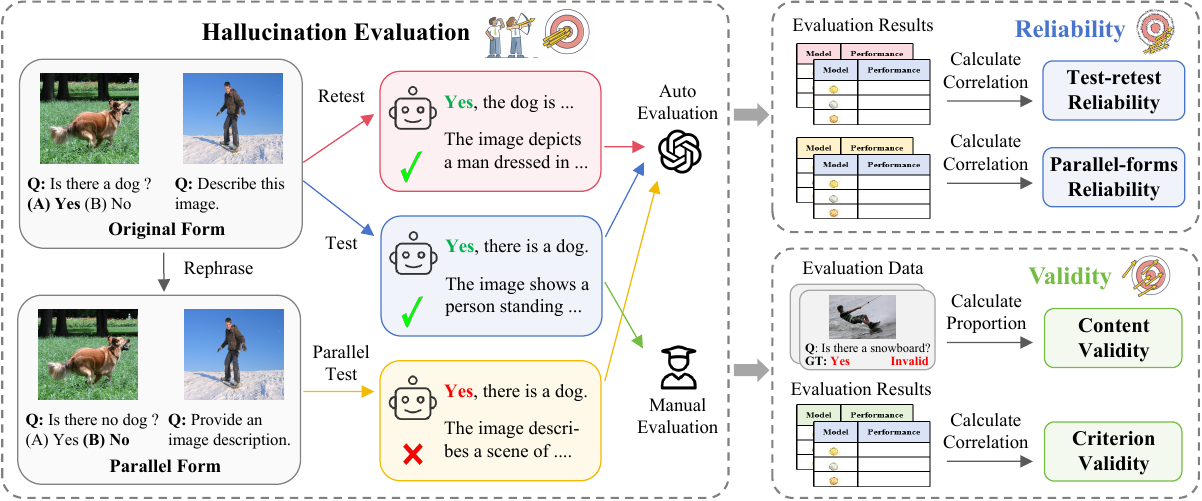}
    \caption{Overview of our Hallucination benchmark Quality Measurement framework (HQM). For reliability, we examine whether the \includegraphics[height=0.90\baselineskip]{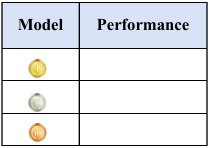} auto evaluation test results are consistent with \includegraphics[height=0.90\baselineskip]{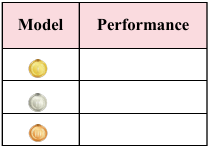} retest results and \includegraphics[height=0.90\baselineskip]{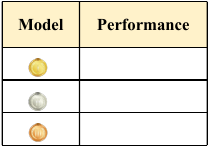} parallel test results. For validity, we measure whether the evaluation data accurately reflects the intended content and whether the \includegraphics[height=0.90\baselineskip]{Figure/logo.pdf} auto evaluation results aligns with \includegraphics[height=0.90\baselineskip]{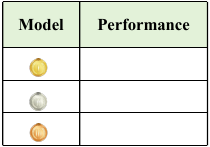} manual evaluation results.}
  \label{overview}
\end{figure*}

To assess hallucination in LVLMs, previous studies~\cite{Li2023pope, wang2023amber, benkish2024openCHAIR, liu2023gaive} have proposed various hallucination benchmarks, supporting evaluation across different tasks. 
However, the quality of these evaluations remains unverified. We observe that some of these benchmarks can be susceptible to potential evaluation issues, such as inconsistent results under repeated or parallel tests, incorrect data annotations, or judgments misaligned with human evaluation, as shown in Figure \ref{issue_examples}. These issues may raise doubts about the trustworthiness of their evaluation results, which are often utilized to assess and develop hallucination mitigation methods. Therefore, it is essential to evaluate the quality of existing hallucination benchmarks, as reliable and valid measures are the foundation for future research.

\IEEEpubidadjcol

Inspired by the systematic test quality assessment methodology in psychometrics~\cite{furr2021psychometrics,rust2014modernpsycho}, we propose a framework for measuring the quality of hallucination benchmarks based on the principles of reliability and validity. An overview of the framework is illustrated in Figure \ref{overview}. For reliability, we assess test-retest reliability and parallel-forms reliability, examining whether the evaluation results are consistent under repeated tests and parallel tests. For validity, we measure content validity and criterion validity, i.e., whether the evaluation data accurately reflects the intended content, and whether the automatic evaluation results align with human evaluation.
Our empirical analysis using this framework reveals and pinpoints certain reliability and validity issues in existing benchmarks. 
We find that while benchmarks built on closed-ended tasks, such as yes-or-no questions and multiple-choice questions, offer efficient automated evaluation, they are prone to the models' inherent response bias~\cite{tjuatja2023llmsresponsebias}, which can lead to inconsistent evaluation results and undermine reliability. Such bias like acquiescence bias in yes-or-no questions (a tendency to always answer "\textit{yes}")~\cite{Li2023pope, fu2023mme} and position bias in multiple-choice questions (a tendency to favor a specific option)~\cite{llm-mcq-bias, mmbench}, could cause models to give answers based on predictable patterns rather than true visual understanding. In contrast, benchmarks based on open-ended tasks like image captioning and free-form VQA, avoid response bias by enabling more flexible responses. But due to the complexity and diversity of the responses, which makes evaluation more challenging, they primarily suffer from validity issues, with a more severe misalignment with human evaluation. Additionally, we also observe that some existing benchmarks contain a certain proportion of annotation noise in their datasets, which also impacts the validity of the evaluation results.

In order to provide a more credible hallucination evaluation, we construct a new hallucination benchmark with improved reliability and validity. We collect images from the Visual Genome~\cite{krishna2017visualgenome} dataset and design image-instruction pairs to evaluate various hallucination types, including object-level, attribute-level, and scene-level hallucinations. 
Due to the inherent influence of yes-or-no biases in closed-ended task settings, we design free-form questions based on the image annotations from Visual Genome covering fine-grained perceptual dimensions, such as object existence and count, attribute properties like color and action, and scene aspects like spatial relation, comparison relation, environment, and text. 
Considering the potential annotation noise, we manually review all image-instruction pairs and remove low-quality samples with inaccurate instructions or incorrect ground truth answers to ensure data quality. 

For evaluation metrics, we note that rule-based syntactic matching metrics, such as accuracy and CHAIR~\cite{rohrbach2018chair}, are often inadequate for comprehensively evaluating hallucinations. In contrast, LLM-based metrics offer greater flexibility for assessing the diverse range of responses in open-ended tasks. However, most of these metrics~\cite{sun2023mmhal, liu2023gaive} rely on external evaluators, such as GPT~\cite{openai2022chatgpt}, to assign specific hallucination scores. Our quality measurement of existing benchmarks~\cite{sun2023mmhal, liu2023gaive} reveals that current LLMs often struggle to assign consistent and accurate scores, aligning with the findings of previous studies~\cite{zheng2023llm-as-a-judge, chen2024mllmasajudge}, which highlight significant discrepancies between model judgments and human preferences in subjective scoring tasks.
To address this issue, we propose a more explicit and objective evaluation approach, which instructs the model to first assess whether the main answer semantically matches the ground truth, and then quantifies any extra hallucinated claims that contradict the image content. We compute the average hallucination rate and the number of hallucination claims as our primary metrics. This approach offers a more factual way to quantify hallucinations compared to subjective scoring and allows us to pinpoint specific sources of hallucination for deeper analysis. Our HQM framework has demonstrated that this proposed metric narrows the evaluation gap between LLMs and human evaluators, thereby improving the reliability and validity of our benchmark.

In conclusion, our contributions are as follows: 

\begin{itemize}[leftmargin=1em]

\item We propose a novel \textbf{H}allucination benchmark \textbf{Q}uality \textbf{M}easurement framework (\textbf{HQM}) for LVLMs, consisting of various indicators to assess both reliability and validity. 

\item Based on our empirical analysis using HQM, we construct a new \textbf{H}igh-\textbf{Q}uality \textbf{H}allucination Benchmark (\textbf{HQH}) with improved reliability and validity, providing a credible evaluation tool.

\item We conduct a large-scale evaluation of representative LVLMs on our HQH, providing in-depth insights into hallucination issues in existing models and informing directions for future improvements.
\end{itemize}

\section{Related Works}

\subsection{Large Vision-Language Models}

Built on the success of LLMs, LVLMs have rapidly developed, demonstrating strong capabilities. Researchers have constructed a series of advanced LVLMs using various methods. CLIP \cite{radford2021clip} pioneers the use of contrastive learning to align images and text, forming a foundation for later models. BLIP2 \cite{li2023blip} adopts a lightweight Q-Former architecture and uses cross-attention mechanisms to align textual and visual representations. InstructBLIP \cite{dai2024instructblip} incorporates textual instructions into the Q-Former, enhancing the model performance. LLaVA \cite{liu2023llava} is the first to introduce instruction tuning techniques to the multimodal field, forming the most mature open-source multimodal model. The emergence of other open-source models such as  Otter~\cite{li2023otter}, Shikra~\cite{chen2023shikra}, and Qwen-VL~\cite{Qwen-VL} have further greatly propelled the development of LVLMs. Additionally, many powerful closed-source LVLMs, including Gemini-1.5-Pro~\cite{team2023gemini} and GPT-4o~\cite{openai2024gpt4o}, have publicly released their APIs, promoting the development of downstream applications. In this paper, we use a wide range of open-source LVLMs as test models under our HQM framework, and benchmark them along with several closed-source models on our HQH.

\subsection{Hallucination Benchmarks for LVLMs}

In the context of LVLMs, hallucination refers to the inconsistency of the generated textual content and the visual input~\cite{bai2024hallucinationsurvey,liu2024survey,zhang2025enhancing}. To evaluate the degree of hallucination in LVLMs, various hallucination benchmarks have been proposed, mainly built upon two categories of tasks, closed-ended and open-ended tasks. For closed-ended tasks, previous works design yes-or-no questions or multiple-choice questions~\cite{lu2023emma}, using accuracy as evaluation metric. For example, POPE~\cite{Li2023pope} constructs yes-or-no questions based on different polling strategies to detect whether the responses contain non-existent objects, primarily focusing on object-level hallucination. Following works like AMBER~\cite{wang2023amber} extend yes-or-no questions to other types of hallucination like attribute-level hallucination. HallusionBench~\cite{guan2023hallusionbench} manually constructs yes-or-no pairs with an innovative structure by human experts, further measuring more fine-grained hallucination. For open-ended tasks, existing works often employ image captioning or free-form VQA. One kind of evaluation metric is CHAIR~\cite{rohrbach2018chair} and its variants~\cite{jing2023faithscore, benkish2024openCHAIR}, which calculates the proportion of hallucinated objects to all objects mentioned in the response and is mostly used for image captioning. For instance, OpenCHAIR~\cite{benkish2024openCHAIR} leverages OCH, which expands CHAIR to an open vocabulary, to evaluate the hallucination in image descriptions. Another kind of metric hallucination score utilizes external LLMs like GPT~\cite{openai2022chatgpt} to exactly grade the hallucination degree of the generated responses, which is relatively more popular in benchmarks built on free-form VQA like MMHal~\cite{sun2023mmhal} and GAVIE~\cite{liu2023gaive}.
We select several representative benchmarks to conduct quality measurement.

\subsection{Benchmark Quality Evaluation}
The development of large models has led to the creation of various benchmarks used to assess their capabilities and risks~\cite{zhang2024b, li2025fakebench, shahreza2025foundation, chen2024we}. However, there are growing concerns about the quality of current AI benchmarks~\cite{mitchell2023we, reuel2024betterbench}. High-quality benchmarks are crucial for ensuring trustworthy evaluations in AI research, as they enable effective model performance comparisons, track progress, identify weaknesses, inform model selection for downstream tasks, and influence policy decisions. Betterbench~\cite{reuel2024betterbench} develops an AI benchmark assessment framework based on 46 criteria derived from expert interviews and domain literature. In contrast to Betterbench, which focuses on benchmark design and usability, our HQM, guided by psychometric theory, emphasizes the reliability and validity of benchmark evaluation, proposing corresponding quantitative metrics. We believe this framework will help uncover potential reliability and validity issues in existing benchmarks and inspire improvements from a new perspective.

\section{Hallucination Benchmark Quality Measurement Framework}
\label{HQM}

Psychometrics, the science of psychological assessment, has long been used to measure constructs such as human-like intelligence~\cite{rust2014modernpsycho}. Functionally, AI benchmarks for evaluating model capabilities share many similarities with psychological tests, including common concerns such as test quality. Therefore, integrating psychometric principles into AI evaluation has received increasing attention~\cite{wang2023aipsyco, pellert2023aipsycho, ye2025measuring}. Inspired by the systematic methodology for assessing psychological test quality, we propose a hallucination benchmark quality evaluation framework, HQM, which comprises multiple indicators focusing on both reliability and validity of hallucination benchmarks. An overview of the HQM framework is shown in Figure~\ref{overview}.

\subsection{Reliability}
\label{realibity}

Reliability refers to the consistency or stability of a test~\cite{rust2014modernpsycho, wang2023aipsyco}. We leverage two reliability indicators, test-retest reliability and parallel-forms reliability~\cite{rust2014modernpsycho}, to quantify the reliability of a hallucination benchmark.

\textbf{Test-retest Reliability.} We use test-retest reliability to reflect the consistency of evaluation results under repeated tests, also known as replicability. Specifically, for each benchmark, we conduct two repeated tests on the same set of test models with different random seeds. The Pearson correlation coefficient~\cite{galton1877pearson} between the two sets of results is calculated as the test-retest reliability:
\begin{equation}
\textit{Test-retest Reliability} = \frac{Cov(S, S_{retest})}{\sigma_{S}  \sigma_{S_{retest}}},
\end{equation}
where $S$ represents the original test results, $S_{retest}$ represents the retest results, $Cov$ denotes covariance, and $\sigma$ denotes standard deviation. We expect the two sets of results to be at a consistent level, without significant fluctuations. Higher test-retest reliability indicates that the benchmark is less sensitive to random variations during the evaluation process, such as those caused by different random seeds or introduced by external tools used in evaluation.

\textbf{Parallel-forms Reliability.}
Parallel-forms reliability is utilized to illustrate the consistency of evaluation results across parallel tests, which is somewhat analogous to robustness. For each benchmark, we create a parallel version by generating equivalent prompts, i.e., rephrasing the instructions in alternative prompt formats. In detail, yes-or-no questions are rewritten into negotiated forms with the opposite ground truth answers, the order of options in multiple-choice questions is randomly shuffled after rephrasing, and captioning instructions and free-form questions are rephrased using synonymous expressions. 
Similar to test-retest reliability, we test the two parallel-forms benchmarks on the same models and calculate their correlation coefficient:
\begin{equation}
\textit{Parallel-forms Reliability} = \frac{Cov(S, S_{parallel})}{\sigma_{S} \sigma_{S_{parallel}}}
\end{equation}
where $S$ represents the original evaluation results and $S_{parallel}$ represents the results of the parallel form. Higher parallel-forms reliability suggests that the benchmark is less influenced by the response bias introduced by specific task prompt.

\begin{table*}[t]
\centering
\caption{Quality measurement results of hallucination benchmarks. The upper benchmarks are based on closed-ended tasks, while the lower benchmarks are built on open-ended tasks. \textbf{Hal} denotes hallucination. The top-2 results are \textbf{bolded} and \underline{underlined}, respectively.}
\label{qualitytable}
\begin{adjustbox}{max width=\linewidth}
\begin{tabular}{l c c | c c c | c c | c c}
\toprule
 \multicolumn{3}{c|}{\textbf{Benchmark Details}} &  \multicolumn{3}{c|}{\textbf{Hallucination Type}} & \multicolumn{2}{c|}{\textbf{Reliability}} & \multicolumn{2}{c}{\textbf{Validity}}\\ 
\cmidrule(lr){1-3} \cmidrule(lr){4-6} \cmidrule(lr){7-8} \cmidrule(lr){9-10} 
\textbf{Name} & \textbf{Task} & \textbf{Metric} &\textbf{Object-level} &\textbf{Attribute-level} &\textbf{Scene-level} & \textbf{Test-retest} & \textbf{Parallel-forms} & \textbf{Content} & \textbf{Criterion} \\

\midrule
POPE & Yes-or-No & Accuracy & \vmark & \xmark & \xmark & \textbf{0.9996} & 0.3563 & 0.84 & \textbf{0.9634} \\ 
AMBER-d & Yes-or-No & Accuracy & \vmark & \vmark & \vmark & 0.9986 & 0.3636 & 0.90 & 0.9321 \\ 
\midrule
AMBER-g & Captioning & CHAIR & \vmark & \xmark & \xmark & 0.9378 & 0.5333 & 0.74 & 0.8774 \\ 
OpenCHAIR & Captioning & OCH & \vmark & \xmark & \xmark & 0.9896 & 0.5510 & 0.76& 0.6818 \\ 
MMHal & Free-form & Hal Score & \vmark & \vmark & \vmark & 0.8829 & 0.8833 & 0.79& 0.7518 \\ 
GAVIE & Free-form & Hal Score & \vmark & \vmark & \vmark & 0.9091 & 0.9136 &0.88 & 0.7938 \\ 
HQH(Ours) & Free-form & Hal Rate & \vmark & \vmark & \vmark & 0.9977 & \textbf{0.9856} & - & \underline{0.9556} \\ 
\bottomrule
\end{tabular}
\end{adjustbox}
\end{table*}

\subsection{Validity}
\label{validity}

Validity indicates how well a test measures what it is designed to measure~\cite{rust2014modernpsycho}.  To assess the validity of a hallucination benchmark, we leverage the content validity and criterion validity~\cite{whitely1983construct, brod2009qualitative}.

\textbf{Content Validity.} Content validity assesses whether the evaluation data appropriately and effectively reflects the intended content that the benchmark is designed to evaluate. For hallucination benchmarks, we manually review and verify whether each image-instruction pair accurately matches the correct ground truth answer and aligns with the intended hallucination types. We quantify content validity by calculating the proportion of valid samples:
\begin{equation}
\textit{Content Validity} = \frac{N_{valid}}{N},
\end{equation}
where $N_{valid}$ represents the number of valid samples and $N$ denotes the total number of samples. Ideally, a hallucination benchmark should ensure that the data is free from annotation errors, thereby supporting an accurate analysis of model performance across different hallucination types.

\textbf{Criterion Validity.} Criterion validity measures the extent to which evaluation results correlate with a trusted criterion. 
We conduct human evaluation on the responses of different models for each hallucination benchmark, identifying whether they contain content inconsistent with the input image, and using the human evaluation results as a reliable criterion reference.
The criterion validity is then quantified by the correlation between the automated benchmark evaluation results and human evaluation results:
\begin{equation}
\textit{Criterion Validity} = \frac{Cov(S, S_{human})}{\sigma_{S} \sigma_{S_{human}}},
\end{equation}
where $S$ represents the auto benchmark evaluation results and $S_{human}$ represents human evaluation results. Higher criterion validity illustrates the evaluation metric is more accurate and effective.

Since the calculation of both content and criterion validity requires the participation of human annotators, which is resource-intensive, for efficiency, we randomly sample a subset of image-instruction pairs from each benchmark for validity measurement. 

\subsection{Quality Measurement}
\label{qualityresults}

We select 6 representative publicly available hallucination benchmarks, POPE~\cite{Li2023pope}, AMBER (including AMBER-d and AMBER-g)~\cite{wang2023amber}, 
OpenCHAIR~\cite{benkish2024openCHAIR}, MMHal~\cite{sun2023mmhal}, and GAVIE~\cite{liu2023gaive}, for quality measurement. Regarding the evaluation metrics, POPE and AMBER-d
use accuracy on Yes-or-No questions, AMBER-g employs CHAIR, which calculates the proportion of hallucinated objects in the image descriptions based on all mentioned objects, OpenCHAIR uses OCH, which expands CHAIR to support an open vocabulary, MMHal and GAVIE adopt hallucination score, leveraging external LLM to assess the degree of hallucination in model responses. Benchmarks requiring external LLM assistance are conducted using GPT-4o~\cite{openai2024gpt4o}. We test on 9 currently mainstream open-source LVLMs, BLIP2~\cite{li2023blip}, InstructBLIP~\cite{dai2024instructblip}, InternLM-Xcomposer-VL~\cite{zhang2023internlm}, LLaVA-1.5~\cite{liu2023llava}, MiniGPT4~\cite{zhu2023minigpt}, MiniGPT-V2~\cite{chen2023minigptv2}, Otter~\cite{li2023otter}, Qwen-VL~\cite{Qwen-VL} and Shikra~\cite{chen2023shikra}, with a total of 20 checkpoints.

\begin{table}[t!]
\centering
\caption{Partial evaluation results on POPE and AMBER-d under original test and parallel test. \textbf{Acc} denotes the accuracy. \textbf{Yes\%} denotes the proportion of responses answering "\textit{yes}" to the given question. \textbf{-p} denotes the results under parallel test.}
\begin{adjustbox}{max width=\linewidth}
\begin{tabular}{l c c c c c}
\toprule
\textbf{ \multirow{2}{*}[-0.5ex]{Model}} & \multicolumn{2}{c}{\textbf{POPE}} & \multicolumn{2}{c}{\textbf{POPE-p}} \\
\cmidrule(lr){2-3} \cmidrule(lr){4-5} 
 & \textbf{\hspace{0.5em}Acc $\uparrow$\hspace{0.5em}} & \textbf{\hspace{0.5em}Yes\%\hspace{0.5em}} & \textbf{\hspace{0.5em}Acc $\uparrow$\hspace{0.5em}} & \textbf{\hspace{0.5em}Yes\%\hspace{0.5em}} \\
 \midrule
MiniGPT4-Llama2\hspace{4.15em} & 0.548 & 0.883 & 0.463 & 0.818 \\
Otter & 0.661 & 0.759 & 0.461 & 0.804 \\
MiniGPT4-Vicuna-7B & 0.548 & 0.202 & 0.497 & 0.184 \\
Qwen-VL-7B & 0.791 & 0.325 & 0.500 & 0.021 \\
 \midrule
\textbf{ \multirow{2}{*}[-0.5ex]{Model}} & \multicolumn{2}{c}{\textbf{AMBER-d}} & \multicolumn{2}{c}{\textbf{AMBER-d-p}} \\
\cmidrule(lr){2-3} \cmidrule(lr){4-5} 
 & \textbf{Acc $\uparrow$} & \textbf{Yes\%} & \textbf{Acc $\uparrow$} & \textbf{Yes\%} \\
 \midrule
MiniGPT4-Llama2 & 0.461 & 0.783 & 0.538 & 0.706 \\
Otter & 0.595 & 0.715 & 0.438 & 0.756 \\
MiniGPT4-Vicuna-7B & 0.622 & 0.251 & 0.398 & 0.217 \\
Qwen-VL-7B & 0.761 & 0.193 & 0.440 & 0.220 \\

\bottomrule
\end{tabular}
\end{adjustbox}
\label{yesratio}
\end{table}

Table \ref{qualitytable} presents the overall quality measurement results under our HQM framework. In general, benchmarks built on open-ended tasks show superior reliability, while those based on closed-ended tasks exhibit stronger validity. Specifically, in terms of test-retest reliability, free-form VQA-based benchmarks exhibit slightly lower performance, primarily due to the use of external LLM judge, which introduces a certain degree of external randomness in the hallucination scoring process. Previous studies highlight that using LLMs as scoring evaluators may lead to unstable results~\cite{chen2024mllmasajudge}, contributing to inconsistent scores to similar model responses. 

In contrast, regarding parallel-forms reliability, benchmarks with closed-ended tasks reveal obvious shortcomings due to the response bias of models towards specific task prompts, such as acquiescence bias and dissent bias. In the evaluation of POPE and AMBER-d, 
we calculate the yes-ratio of each model, which is defined as the proportion of model responses answering "\textit{yes}" to the given yes-or-no questions. As shown in Table \ref{yesratio}, we find that MiniGPT4-Llama2 and Otter suffer from significant acquiescence bias, exhibiting a strong tendency to answer "\textit{yes}" regardless of question content, with substantially high yes-ratios. Meanwhile, MiniGPT4-Vicuna-7B and Qwen-VL-7B encounter great dissent bias, characterized by a persistent inclination to respond "\textit{no}", leading to apparently low yes-ratios. Such inherent bias causes great performance gap in parallel tests, making it unclear whether the evaluation result truly reflects models' hallucination level or is simply an artifact of their own response bias. Likewise, the parallel-forms reliability of captioning-based benchmarks, AMBER-g and OpenCHAIR, is also unsatisfactory. This is because, in captioning tasks, the response lengths of certain models are significantly influenced by the design of prompt. As shown in Table \ref{avglen}, given equivalent prompts, "\textit{Describe the image.}" in original test and "\textit{Provide a description of the image.}" in parallel tests, the average response lengths of some models fluctuate greatly. Since longer responses are empirically more prone to hallucinations, the variation in response length caused by changes in prompt format further leads to inconsistencies in hallucination evaluation results under parallel tests.

\begin{figure}
  \centering
  \includegraphics[width=\linewidth]{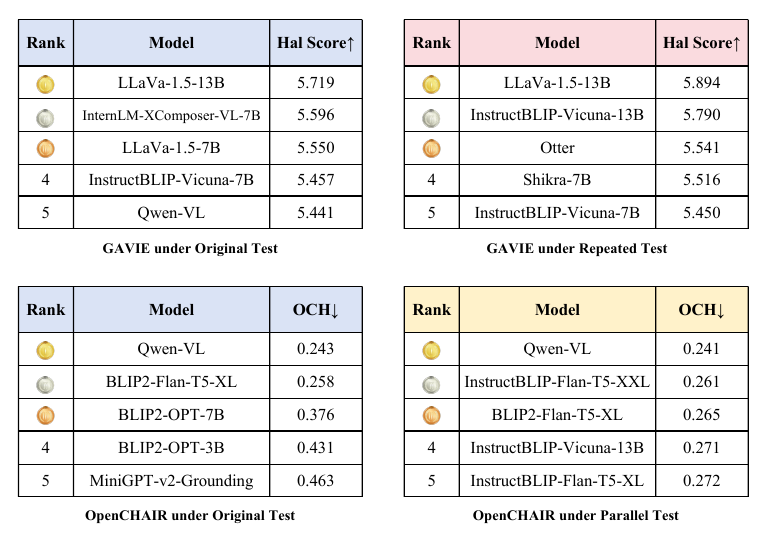}
  \caption{Partial leaderboards of evaluated open-source LVLMs across various tests in quality measurement.}
  \label{leaderboard}
\end{figure}

Some leaderboards for these benchmarks under repeated and parallel tests are shown in Figure \ref{leaderboard}. As illustrated, there are significant discrepancies in model performance and rankings across the original, repeated, and parallel tests, which raises concerns about the reliability of the evaluation.

\begin{table}[t!]
\centering
\caption{Partial evaluation results on AMBER-g and OpenCHAIR under original test and parallel test. \textbf{Len} denotes the average length of model responses, i.e., average word counts. \textbf{-p} denotes the results under parallel test.}
\begin{adjustbox}{max width=\linewidth}
\begin{tabular}{l c c c c c c c c}
\toprule
\textbf{ \multirow{2}{*}[-0.5ex]{Model}} &\multicolumn{2}{c}{\textbf{AMBER-g}} &\multicolumn{2}{c}{\textbf{AMBER-g-p}} \\
\cmidrule(lr){2-3} \cmidrule(lr){4-5} 
 &  \textbf{CHAIR $\downarrow$} &	\textbf{Len} &  \textbf{CHAIR $\downarrow$} &	\textbf{Len}  \\
\midrule
InstructBLIP-Flan-T5-XXL &0.151&104.66 &0.037&10.37 \\
InstructBLIP-Vicuna-7B &0.085&80.53 &0.031&10.66 \\
InternLM-XComposer-VL-7B & 0.109 & 56.44 & 0.044 & 22.53 \\
Otter &0.102&47.15 &0.128&63.46 \\
\midrule
\textbf{ \multirow{2}{*}[-0.5ex]{Model}} &\multicolumn{2}{c}{\textbf{OpenCHAIR}} &\multicolumn{2}{c}{\textbf{OpenCHAIR-p}} \\
\cmidrule(lr){2-3} \cmidrule(lr){4-5} 
 &  \textbf{OCH $\downarrow$} &	\textbf{Len} &  \textbf{OCH $\downarrow$} &	\textbf{Len}  \\
\midrule
InstructBLIP-Flan-T5-XXL &0.525 &103.07 &0.261 &10.29 \\
InstructBLIP-Vicuna-7B &0.470 &93.04 &0.265 &10.85 \\
InternLM-XComposer-VL-7B & 0.470 & 64.33 & 0.433 & 25.91 \\
Otter &0.493 &55.94 &0.506 &63.69 \\
\bottomrule
\end{tabular}
\end{adjustbox}

\label{avglen}
\end{table}

For content validity, we quantify the proportion of valid content to measure the quality of evaluation data. Specifically, for yes-or-no and free-form questions, we primarily check whether the instructions and their corresponding standard answers or annotations are accurate, clear, and unambiguous, and whether the hallucination type label, if provided, aligns with the intended hallucination type to be detected. For captioning tasks, we focus on identifying errors in the ground truth caption or omissions of key objects. Since most existing benchmarks are constructed based on annotations from existing datasets such as MSCOCO~\cite{lin2014mscoco}, which are known contain annotation errors~\cite{schubert2024labelerror, neuhaus2025repope}, we find that they suffer from a substantial degree of invalid content. For instance, POPE and AMBER-d, which are built upon MSCOCO annotations, exhibit over 10\% invalid content, primarily due to noise in the original dataset, which significantly impacts the validity of the results. This highlights the need to avoid over-relying on annotations from existing datasets when constructing hallucination benchmarks, and emphasizes the importance of filtering the data after construction.

Except for content validity, we investigate criterion validity. Although closed-ended tasks restrict models to choosing from a given set of potential answers, their evaluation is still not completely aligned with human evaluation. This discrepancy arises because some models do not strictly follow the prompt to generate only the given form of answers such as "\textit{yes}", "\textit{no}" or options "\textit{A, B, C}", instead, they may append their own analysis after providing their choice. A common occurrence during the evaluation is that the model provides the correct answer, but there exists hallucination in the analysis which is contradictory to the answer. Meanwhile, the evaluation of open-ended tasks encounters more significant criterion validity issues. In captioning-based benchmarks AMBER-g and OpenCHAIR, both metrics calculate only the proportion of hallucinated objects among all mentioned objects, detecting only existence hallucination. This results in the misalignment with human evaluation since image descriptions usually contain multiple types of hallucination. In free-form VQA, hallucination score, which leverages external LLM to assign a specific score to the hallucination level of model response, also presents limitations. The main reason, in our view, is that it is too difficult for current LLMs to consistently and accurately grade the degree of hallucination in model responses.
Even with provided specific guidelines, there remains a gap between LLMs and human in scoring evaluation~\cite{zheng2023llm-as-a-judge, chen2024mllmasajudge}.

\begin{figure*}[t]
  \begin{center}
\includegraphics[width=\linewidth]{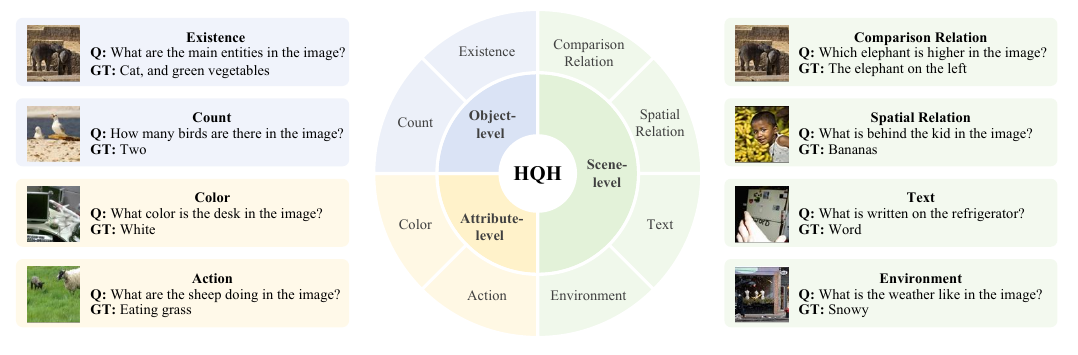}
  \end{center}
  \caption{Examples of image-instruction pairs for different hallucination evaluation dimension in HQH.}
  \label{hallucinationtype}
\end{figure*}

\begin{figure*}[t]
  \begin{center}
  \includegraphics[width=0.95\linewidth]{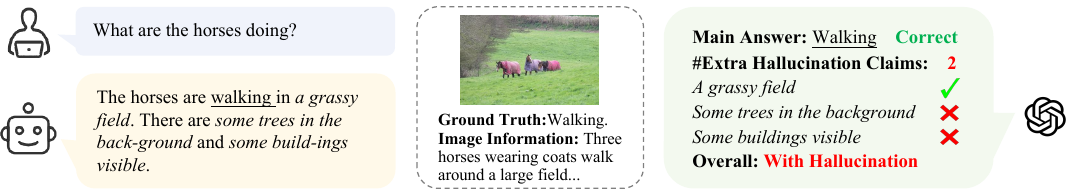}
  \end{center}
  \caption{Illustration of HQH evaluation.}

\label{evaluation_illustration}
\end{figure*}

\section{High-Quality Hallucination Benchmark}
Our empirical analysis using HQM reveals and pinpoints potential reliability and validity issues in existing benchmarks, highlighting a critical gap in current hallucination evaluation. To address this, we propose HQH, a high-quality hallucination benchmark, which demonstrates improved reliability and validity under HQM, offering a more credible evaluation tool.

\subsection{Data Construction}

Considering that closed-ended settings inevitably introduce response bias to some models as illustrated in Section \ref{qualityresults}, our HQH is built on open-ended tasks. Since the evaluation results of captioning tasks fluctuate significantly with different prompts, leading to reliability issues, and given the difficulty of comprehensively assessing different types of hallucinations through captioning tasks, we opt to conduct our evaluation using free-form VQA.

We collect images from the Visual Genome~\cite{krishna2017visualgenome} dataset. Based on the image annotations in the dataset, we leverage advanced LLM to extract the key information and generate candidate questions along with their corresponding ground truth answers for object-level, attribute-level, and scene-level hallucination evaluation. These questions cover 8 fine-grained visual perceptual dimensions, including object existence and count, attribute properties like color and action, and scene aspects like spatial relation, comparison relation, environment, and text.

To address potential annotation noise and ensure data quality, we conduct a manual review of all candidate image-instruction pairs, removing low-quality samples where instructions are inaccurate (e.g., ambiguous subject reference) or where the ground truth answers are incorrect (e.g., misaligned with the image).
The constructed HQH consists of 4000 image-instruction pairs, with 500 pairs for each dimension. Examples are shown in Figure \ref{hallucinationtype}.

\begin{table*}[t]
\centering
\caption{Evaluation results of LVLMs on HQH. \textbf{Hal\%} denotes the average hallucination rate, \textbf{\#Hal} denotes the average number of hallucination claims. The top-2 results are \textbf{bolded} and \underline{underlined}, respectively.}
\begin{adjustbox}{max width=\linewidth}
\setlength{\tabcolsep}{12pt}
\begin{tabular}{@{}lccccccc@{}}
\toprule
\textbf{ \multirow{2}{*}[-0.5ex]{Model}} & \multicolumn{4}{c}{\textbf{Main Hal\% $\downarrow$}} & \textbf{ \multirow{2}{*}[-0.5ex]{Extra \#Hal $\downarrow$}} & \textbf{ \multirow{2}{*}[-0.5ex]{Overall Hal\% $\downarrow$}} \\
\cmidrule(lr){2-5}
 & \textbf{Object-level} & \textbf{Attribute-level} & \textbf{Scene-level} & \textbf{Average} & & \\

\midrule
BLIP2-OPT-3B             & \cellcolor[HTML]{FAFBFD}0.806          & \cellcolor[HTML]{F5F7FC}0.714          & \cellcolor[HTML]{FFFFFF}0.886          & \cellcolor[HTML]{FBFCFD}0.823          & \cellcolor[HTML]{FFF9E4}0.408          & \cellcolor[HTML]{FFFFFF}0.863          \\
BLIP2-OPT-7B             & \cellcolor[HTML]{FAFBFD}0.808          & \cellcolor[HTML]{F1F4FB}0.647          & \cellcolor[HTML]{FEFEFE}0.875          & \cellcolor[HTML]{FAFBFD}0.801          & \cellcolor[HTML]{FFF9E6}0.431          & \cellcolor[HTML]{FCFEFC}0.826          \\
MiniGPT4-Vicuna-7B       & \cellcolor[HTML]{F3F6FB}0.694          & \cellcolor[HTML]{F0F3FA}0.635          & \cellcolor[HTML]{FDFEFE}0.866          & \cellcolor[HTML]{F8F9FC}0.765          & \cellcolor[HTML]{FFFEFB}0.728          & \cellcolor[HTML]{FDFEFC}0.832          \\
MiniGPT4-Vicuna-13B      & \cellcolor[HTML]{F5F7FC}0.720          & \cellcolor[HTML]{EDF1FA}0.590          & \cellcolor[HTML]{FDFDFE}0.851          & \cellcolor[HTML]{F7F8FC}0.753          & \cellcolor[HTML]{FFFFFF}0.779          & \cellcolor[HTML]{FCFEFC}0.827          \\
MiniGPT4-Llama2          & \cellcolor[HTML]{FBFCFD}0.823          & \cellcolor[HTML]{EDF1FA}0.589          & \cellcolor[HTML]{F9FBFD}0.798          & \cellcolor[HTML]{F7F8FC}0.752          & \cellcolor[HTML]{FFFEFC}0.744          & \cellcolor[HTML]{FCFDFB}0.816          \\
MiniGPT-V2               & \cellcolor[HTML]{F0F3FA}0.642          & \cellcolor[HTML]{EAEEF8}0.522          & \cellcolor[HTML]{F8F9FD}0.767          & \cellcolor[HTML]{F2F5FB}0.675          & \cellcolor[HTML]{FFFDF8}0.684          & \cellcolor[HTML]{F9FCF7}0.760          \\
MiniGPT-V2-VQA           & \cellcolor[HTML]{F1F4FA}0.645          & \cellcolor[HTML]{E5EBF7}0.446          & \cellcolor[HTML]{F4F6FB}0.697          & \cellcolor[HTML]{EFF2FA}0.622          & \cellcolor[HTML]{FFF3CA}{\ul 0.035}    & \cellcolor[HTML]{F1F8EC}0.622          \\
Otter                    & \cellcolor[HTML]{F3F5FB}0.685          & \cellcolor[HTML]{E8EDF8}0.494          & \cellcolor[HTML]{F8F9FC}0.766          & \cellcolor[HTML]{F3F5FB}0.678          & \cellcolor[HTML]{FFF8E3}0.394          & \cellcolor[HTML]{F8FCF6}0.749          \\
BLIP2-Flan-T5-XL         & \cellcolor[HTML]{EEF1FA}0.596          & \cellcolor[HTML]{EBEFF9}0.540          & \cellcolor[HTML]{F9FAFD}0.787          & \cellcolor[HTML]{F3F5FB}0.678          & \cellcolor[HTML]{FFF5D5}0.189          & \cellcolor[HTML]{F5FAF1}0.691          \\
InstructBLIP-Vicuna-7B   & \cellcolor[HTML]{E8EDF8}0.502          & \cellcolor[HTML]{E1E7F6}0.379          & \cellcolor[HTML]{F3F5FB}0.678          & \cellcolor[HTML]{ECF0F9}0.559          & \cellcolor[HTML]{FFF7DD}0.301          & \cellcolor[HTML]{F0F8EB}0.610          \\

InstructBLIP-Vicuna-13B  & \cellcolor[HTML]{EFF2FA}0.609          & \cellcolor[HTML]{E2E8F6}0.392          & \cellcolor[HTML]{F4F6FC}0.708          & \cellcolor[HTML]{EEF2FA}0.604          & \cellcolor[HTML]{FFF8E3}0.389          & \cellcolor[HTML]{F3F9EF}0.662          \\
Shikra-7B                & \cellcolor[HTML]{E7ECF8}0.474          & \cellcolor[HTML]{E2E8F6}0.391          & \cellcolor[HTML]{F2F4FB}0.663          & \cellcolor[HTML]{EBEFF9}0.548          & \cellcolor[HTML]{FFF7DF}0.332          & \cellcolor[HTML]{F2F8ED}0.634          \\
Shikra-7B-VQA            & \cellcolor[HTML]{EAEEF8}0.525          & \cellcolor[HTML]{DFE6F5}0.342          & \cellcolor[HTML]{ECF0F9}0.559          & \cellcolor[HTML]{E8EDF8}0.496          & \cellcolor[HTML]{FFF3CA}\textbf{0.032} & \cellcolor[HTML]{EAF5E2}0.496          \\
InternLM-XComposer-VL-7B & \cellcolor[HTML]{EBEFF9}0.539          & \cellcolor[HTML]{E3E9F7}0.414          & \cellcolor[HTML]{F4F6FB}0.704          & \cellcolor[HTML]{EDF1FA}0.590          & \cellcolor[HTML]{FFF4CF}0.109          & \cellcolor[HTML]{EFF8EA}0.599          \\
InstructBLIP-Flan-T5-XL  & \cellcolor[HTML]{E9EDF8}0.511          & \cellcolor[HTML]{E2E8F6}0.382          & \cellcolor[HTML]{F4F6FB}0.703          & \cellcolor[HTML]{EDF0F9}0.575          & \cellcolor[HTML]{FFF5D4}0.180          & \cellcolor[HTML]{EFF7EA}0.594          \\
InstructBLIP-Flan-T5-XXL & \cellcolor[HTML]{E7ECF8}0.483          & \cellcolor[HTML]{E2E8F6}0.398          & \cellcolor[HTML]{F4F6FC}0.708          & \cellcolor[HTML]{EDF0F9}0.574          & \cellcolor[HTML]{FFF5D4}0.175          & \cellcolor[HTML]{EFF7EA}0.596          \\
Yi-VL                    & \cellcolor[HTML]{EAEEF9}0.533          & \cellcolor[HTML]{E4E9F7}0.422          & \cellcolor[HTML]{F0F3FA}0.635          & \cellcolor[HTML]{ECF0F9}0.556          & \cellcolor[HTML]{FFF5D3}0.164          & \cellcolor[HTML]{EFF7E9}0.584          \\
LLaVA-1.5-7B             & \cellcolor[HTML]{EAEFF9}0.536          & \cellcolor[HTML]{E2E8F6}0.390          & \cellcolor[HTML]{EDF1F9}0.579          & \cellcolor[HTML]{EAEEF8}0.521          & \cellcolor[HTML]{FFF6DA}0.271          & \cellcolor[HTML]{EEF7E8}0.574          \\
LLaVA-1.5-13B            & \cellcolor[HTML]{E9EDF8}0.509          & \cellcolor[HTML]{E1E7F6}0.367          & \cellcolor[HTML]{ECF0F9}0.558          & \cellcolor[HTML]{E8EDF8}0.498          & \cellcolor[HTML]{FFF6D9}0.248          & \cellcolor[HTML]{EDF6E6}0.550          \\
mPLUG-Owl2               & \cellcolor[HTML]{E9EDF8}0.508          & \cellcolor[HTML]{E0E7F6}0.364          & \cellcolor[HTML]{EEF1FA}0.598          & \cellcolor[HTML]{E9EEF8}0.517          & \cellcolor[HTML]{FFF5D6}0.211          & \cellcolor[HTML]{EDF7E7}0.563          \\
InternVL2-8B             & \cellcolor[HTML]{E9EDF8}0.507          & \cellcolor[HTML]{E0E6F6}0.348          & \cellcolor[HTML]{EBEFF9}0.549          & \cellcolor[HTML]{E8ECF8}0.488          & \cellcolor[HTML]{FFF7DF}0.333          & \cellcolor[HTML]{EDF6E6}0.547          \\
MiniCPM-Llama2-v2.5      & \cellcolor[HTML]{E2E8F6}0.383          & \cellcolor[HTML]{DCE4F5}0.294          & \cellcolor[HTML]{E4E9F7}0.421          & \cellcolor[HTML]{E1E7F6}0.380          & \cellcolor[HTML]{FFF8E0}0.349          & \cellcolor[HTML]{EBF6E4}0.525          \\
Emu2-CHat                & \cellcolor[HTML]{E0E6F6}0.349          & \cellcolor[HTML]{E0E6F6}0.356          & \cellcolor[HTML]{EEF1FA}0.594          & \cellcolor[HTML]{E7ECF8}0.473          & \cellcolor[HTML]{FFF4D0}0.123          & \cellcolor[HTML]{EAF5E3}0.501          \\
Gemini-1.5-Pro           & \cellcolor[HTML]{E3E9F7}0.406          & \cellcolor[HTML]{DFE5F5}0.334          & \cellcolor[HTML]{E9EEF8}0.512          & \cellcolor[HTML]{E5EAF7}0.441          & \cellcolor[HTML]{FFF6D8}0.231          & \cellcolor[HTML]{E9F4E1}0.480          \\
Phi-3-Vision             & \cellcolor[HTML]{E2E8F6}0.398          & \cellcolor[HTML]{DCE3F5}0.292          & \cellcolor[HTML]{E5EAF7}0.435          & \cellcolor[HTML]{E2E8F6}0.390          & \cellcolor[HTML]{FFF4D2}0.153          & \cellcolor[HTML]{E6F3DD}0.425          \\
GLM-4V-9B                & \cellcolor[HTML]{DEE5F5}0.327          & \cellcolor[HTML]{DAE2F4}0.250          & \cellcolor[HTML]{E0E6F6}\textbf{0.354} & \cellcolor[HTML]{DEE5F5}{\ul 0.321}    & \cellcolor[HTML]{FFF9E6}0.436          & \cellcolor[HTML]{E5F3DD}0.422          \\
Qwen-VL-7B               & \cellcolor[HTML]{EAEEF9}0.526          & \cellcolor[HTML]{DDE4F5}0.306          & \cellcolor[HTML]{E2E8F6}0.386          & \cellcolor[HTML]{E3E8F6}0.401          & \cellcolor[HTML]{FFF3CA}0.039          & \cellcolor[HTML]{E4F2DB}0.403          \\
Qwen2-VL-7B              & \cellcolor[HTML]{DEE5F5}0.329          & \cellcolor[HTML]{D9E1F4}{\ul 0.227}    & \cellcolor[HTML]{E0E7F6}{\ul 0.363}    & \cellcolor[HTML]{DEE5F5}\textbf{0.320} & \cellcolor[HTML]{FFF6D9}0.246          & \cellcolor[HTML]{E4F2DB}0.400          \\
LLaVA-OneVision-7B       & \cellcolor[HTML]{DDE4F5}\textbf{0.304} & \cellcolor[HTML]{DDE4F5}0.310          & \cellcolor[HTML]{E3E9F6}0.403          & \cellcolor[HTML]{E0E6F6}0.355          & \cellcolor[HTML]{FFF4D0}0.122          & \cellcolor[HTML]{E4F2DA}{\ul 0.389}    \\
GPT-4o                   & \cellcolor[HTML]{DDE4F5}{\ul 0.309}    & \cellcolor[HTML]{D9E1F4}\textbf{0.226} & \cellcolor[HTML]{E2E8F6}0.392          & \cellcolor[HTML]{DEE5F5}0.330          & \cellcolor[HTML]{FFF6D7}0.220          & \cellcolor[HTML]{E3F2D9}\textbf{0.370}
\\

\bottomrule
\end{tabular}
\end{adjustbox}
\label{hqhresults}
\end{table*}

\subsection{Evaluation Metric}

Regarding the evaluation metric, we note that rule-based syntactic matching metrics, such as accuracy or CHAIR~\cite{rohrbach2018chair}, are inadequate for comprehensively capturing hallucinations across the entire model response in different tasks. Therefore, we utilize LLM-based evaluation metrics. Existing LLM-based hallucination evaluations mainly focus on scoring, which requires assigning a hallucination degree to model responses. However, our experiments in quality measurement reveal that the hallucination scores used in existing benchmarks exhibit limitations in both reliability and validity. Previous studies on LLM-as-a-judge also highlight that LLMs, when acting as subjective scoring evaluators, often misalign with human judgment~\cite{zheng2023llm-as-a-judge, chen2024mllmasajudge}. It is beyond the capabilities of current LLMs, to consistently and accurately grade hallucinations in responses to the same level as human evaluators. On one hand, LLMs produce inconsistent scores for similar model responses in repeated or parallel tests, negatively affecting reliability. On the other hand, they often provides inaccurate hallucination scores, undermining validity.

To better utilize the LLM's capabilities as a judge and improve evaluation quality, we refine the LLM-assisted evaluation process. Since directly evaluating the entire response does not allow for pinpointing the source of hallucinations or conducting a deeper analysis, we further decompose the evaluation into a more explicit and objective process. First, we instruct the model to directly assess whether the main answer semantically aligns with the ground truth answer and compute the primary hallucination rate. Given that many models often generate additional analysis or explanations despite being prompted to provide direct answers, we further require it to quantify any extra hallucination claims that contradict the image content and calculate the average number of these extra hallucination claims. An illustration of the evaluation process is shown in Figure ~\ref{evaluation_illustration}. To assess overall performance, we combine both aspects and obtain the overall hallucination rate. Our evaluation metrics are formulated as follows:
\begin{equation}
\resizebox{0.89\linewidth}{!}{$\displaystyle
\begin{aligned}
    \textit{Main Hal\%} &= \frac{1}{N} \sum_{i=1}^N \mathbb{I}(Ans_i \not\equiv GT_i),\\
    \textit{Extra \#Hal} &= \frac{1}{N} \sum_{i=1}^N EH_i,\\
    \textit{Overall Hal\%} &= \frac{1}{N} \sum_{i=1}^N \mathbb{I}\Big(Ans_i \not\equiv GT_i \lor EH_i > 0 \Big). \\
\end{aligned}
$}
\end{equation}
where \(i\) indicates response index, \(N\) denotes total number of responses, \(EH_i\) represents the number of extra hallucination claims in the \(i\)-th response.

We utilize text-only evaluation prompt instead of direct visual access to achieve more accurate results, as supported by previous study~\cite{chen2024mllmasajudge}, which shows that LLMs with detailed image descriptions can outperform MLLMs with visual access in multimodal judging.

\subsection{Evaluation Results}

We measure the quality of our HQH under HQM framework. To ensure a fair comparison, we also use GPT-4o~\cite{openai2024gpt4o} as the external LLM in our evaluation. Table \ref{qualitytable} shows that HQH exhibits the highest reliability among all benchmarks, and its validity is also comparable to that of closed-ended tasks. This demonstrates that HQH provides credible and meaningful hallucination evaluation for LVLMs.

We also conduct an extensive evaluation on HQH with a broader set of mainstream popular LVLMs, including powerful closed-source models, GPT-4o~\cite{openai2024gpt4o} and Gemini-1.5-Pro~\cite{team2023gemini}. Evaluation results are presented in Table \ref{hqhresults}.

\textbf{Overall Performance.} As shown in Table \ref{hqhresults}, closed-source model GPT-4o shows the best overall performance among all the models. Some open-source models, such as LLaVA-OneVision~\cite{li2024LLaVAov} and Qwen2-VL~\cite{wang2024qwen2}, also perform competitively, reaching levels comparable to those of closed-source models, which indicates that their architectural or training design
choices effectively contribute to hallucination mitigation. However, most of the models' overall hallucination rate exceeds 40\%, and even the most advanced GPT-4o still exhibits hallucination in over 35\% of its responses. This indicates that there is still substantial room for improvement in mitigating hallucination in LVLMs. 

\begin{figure}[t]
  \centering
\includegraphics[width=0.9\linewidth]{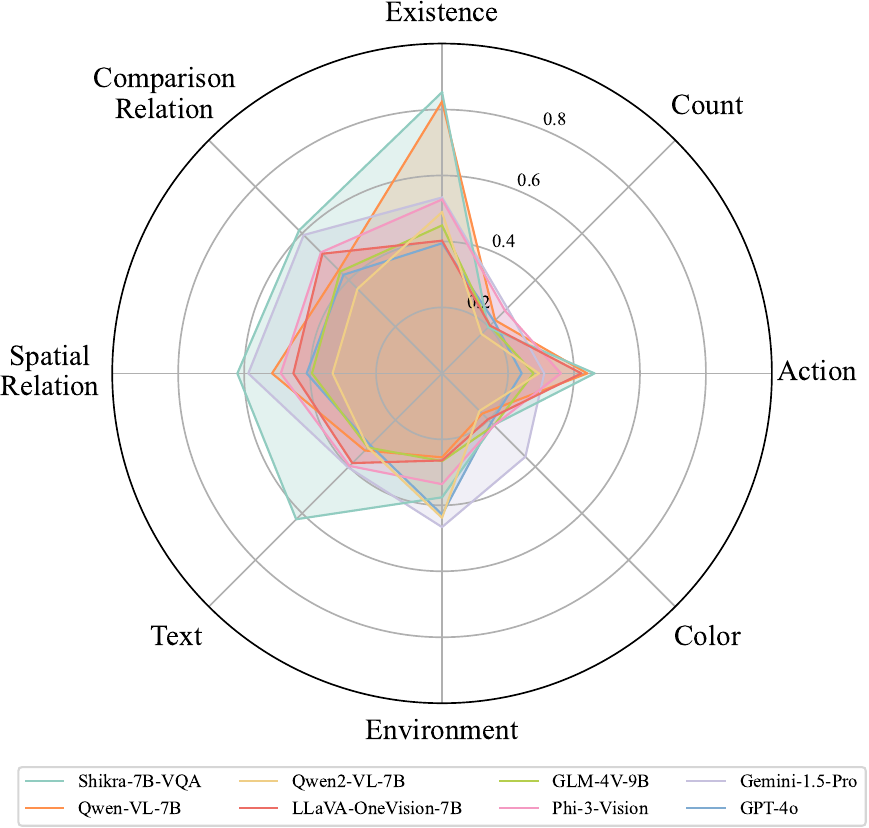}
  \caption{Comparison of main hallucination rates $\downarrow$ of the top-8 LVLMs across different hallucination dimensions. A smaller area indicates better performance.}
  \label{rada}
\end{figure}

\textbf{Hallucination Source.} Upon further analysis, we observe that, in addition to exhibiting hallucinations in the main answers to the questions, many models also demonstrate significant hallucinations in additional content, e.g., explanations, analysis. While these extra hallucinations are often under-recognized, they equally reflect limitations in the models' visual perception capabilities. For instance, models like GLM-4V-9B~\cite{glm2024chatglm4v} perform well in providing main answers to questions, but generate substantial extra hallucinations, revealing deficiencies in their visual processing capabilities. This issue is often overlooked when evaluations focus solely on task accuracy. This highlights that, during hallucination mitigation, it is essential not only to focus on the accuracy of the main answers provided by the model but also to effectively monitor and manage the additional content generated by the model, in order to enhance the model’s practical effectiveness and security comprehensively.

\textbf{Fine-grained Hallucination Dimensions.} We also investigate model performance across different types of hallucinations. To clarify the attribution of hallucinations, we perform a categorical analysis of the main hallucination rates. As shown in Table \ref{hqhresults}, the majority of models exhibit more severe hallucinations at the scene-level, while attribute-level hallucinations are relatively less pronounced. For a more intuitive comparison and in-depth analysis, Figure \ref{rada} provides an comparison of the top-8 LVLMs across fine-grained hallucination dimensions. We find that current LVLMs exhibit comparatively fewer hallucinations related to object count and color attributes, which are more frequently addressed during model training. However, hallucinations related to object existence and relational aspects present more significant challenges, with the average main hallucination rate for object existence reaching 54\%, highlighting the need for greater attention to these issues in future research.

\begin{figure}[t!]
  \centering
  \includegraphics[width=\linewidth]{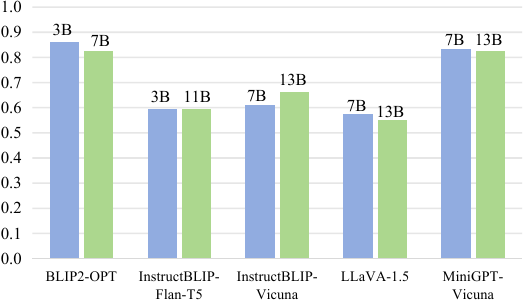}
  \caption{Comparison of the overall hallucination rates $\downarrow$ of LVLMs across different parameter sizes.}
  \label{parameter size}
\end{figure}

\textbf{The Effect of Parameter Size.} We observe that for models of the same architecture, scaling the number of parameters has only a modest impact on hallucination performance. In Figure \ref{parameter size}, we compare the performance of several models with different parameter sizes. While most models show a trend of decreasing hallucination rates as parameter size increases, the reduction is not substantial, with InstructBLIP-Vicuna~\cite{dai2024instructblip} showing nearly identical performance. This suggests that the current scaling of parameters has a limited effect on hallucination reduction, indicating that hallucination mitigation may require improvements in model architecture, training data quality, or task-specific fine-tuning, rather than just increasing parameter size.

\textbf{The Effect of Task Identifier.} Some models, such as MiniGPT-V2~\cite{chen2023minigptv2} and Shikra~\cite{chen2023shikra}, adopt a task-oriented instruction training scheme and provide corresponding task identifier that can be used during inference. For instance, MiniGPT-V2 incorporates the "\textit{[vqa]}" identifier token in its instructions. We observe that adding the VQA task identifier leads to a noticeable reduction in hallucinations, particularly extra hallucinations. This improvement may be attributed to the fact that, without a clear task identifier, the model tends to generate additional, irrelevant information, where the model can easily deviate from the task focus. By introducing task identifiers, the model’s output becomes more focused on the task requirements, thereby reducing such deviations.

\begin{figure}[t!]
  \centering
  \includegraphics[width=\linewidth]{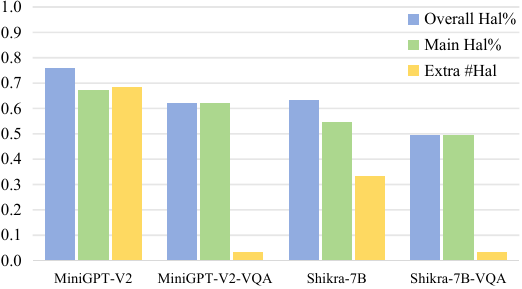}
  \caption{Comparison of model performance of LVLMs with and without task identifier. \textbf{-VQA} denotes with VQA task identifier.}
  \label{task prompt}
\end{figure}

\section{Discussion}

\textbf{AI $\&$ Psychometrics.} Psychometrics is the science of how to maximize the quality of psychological assessments~\cite{rust2014modernpsycho}. As mentioned in Section \ref{HQM}, psychological tests in psychometrics share commonalities with AI evaluation benchmarks. The integration of psychometrics into AI may bring new opportunities for AI evaluation, such as the possibility of using construct-oriented paradigms from psychometrics to evaluate the latent constructs of general AI~\cite{wang2023aipsyco}. Our work focuses mainly on adapting the quality measurement methods of psychological tests to AI benchmarks. There exist other potential combinations that deserve further exploration. 

\textbf{Hallucination Mitigation.} Our work aims to provide a reliable and valid evaluation tool for further hallucination mitigation. Through fine-grained hallucination analysis, we observe that while many models have made progress in addressing common hallucinations, such as those related to attributes, other often overlooked hallucinations like comparison relation hallucinations, remain critical issues. Additionally, some models exhibit significant amounts of extra hallucinations. We recommend implementing targeted optimization during training to address these specific hallucinations, which would help mitigate hallucination issues and enhance the overall robustness of the model.

\section{Conclusion}
We introduce a quality measurement framework for hallucination benchmarks (HQM), utilizing various indicators to assess their reliability and validity. Under our proposed HQM framework, we construct a new high-quality hallucination benchmark (HQH), which is more reliable, valid, and comprehensive. An extensive evaluation of over 15 representative LVLMs, including GPT-4o and Gemini-1.5-Pro, is conducted on our HQH, illustrating that there is still substantial room for improvement. We anticipate that our research will inspire future work to effectively assess and mitigate hallucinations in LVLMs, thereby enhancing model security in practical real-world deployments.

\section{Limitations}

Our proposed HQM framework represents an initial attempt to assess benchmark quality, from the perspective of psychometrics. Since psychometrics is specifically designed for human psychological tests, some measures may not be applicable to AI benchmarks. Currently, we focus on reliability and validity, but in future work, we plan to explore other aspects, such as the internal consistency of benchmarks. Additionally, while our HQM framework is general, it has only been applied to hallucination benchmarks so far. We expect to extend its applications, such as measuring other benchmarks, in the future. Regarding our HQH benchmark, it is built on free-form VQA task. We will continue to expand it by incorporating more tasks to provide a comprehensive evaluation.

\section{Ethics Statement}
Our validity measurement involves human annotators recruited from our institute, manually assessing whether data samples are invalid and model responses exhibit hallucinations. The process does not involve any direct interactions with human participants, and does not have potential risks, such as the collection of identifiable data, exposure to sensitive content, emotional distress, or any other aspects that could impact the participants' rights or well-being. Informed consent is obtained from all participants, and their privacy is strictly protected throughout the study. The entire process adheres to ethical guidelines and has received approval from the Institutional Review Board (IRB).

\bibliographystyle{IEEEtran}
\bibliography{main_abbr}

\end{document}